# A Framework Combining 3D CNN And Transformer For Video-Based Behavior Recognition

Xiuliang Zhang[1], Tadiwa Elisha Nyamasvisva[1], Chuntao Liu[1]

[1](Faculty Of Engineering, Science And Technology, Infrastructure University Kuala Lumpur, Malaysia)

***Abstract:***
*Video-based behavior recognition is essential in fields such as public safety, intelligent surveillance, and human-computer interaction. Traditional 3D Convolutional Neural Network (3D CNN) effectively capture local spatiotemporal features but struggle with modeling long-range dependencies. Conversely, Transformers excel at learning global contextual information but face challenges with high computational costs. To address these limitations, we propose a hybrid framework combining 3D CNN and Transformer architectures. The 3D CNN module extracts low-level spatiotemporal features, while the Transformer module captures long-range temporal dependencies, with a fusion mechanism integrating both representations. Evaluated on benchmark datasets, the proposed model outperforms traditional 3D CNN and standalone Transformers, achieving higher recognition accuracy with manageable complexity. Ablation studies further validate the complementary strengths of the two modules. This hybrid framework offers an effective and scalable solution for video-based behavior recognition..*

## I. Introduction

Video-based behavior recognition has emerged as a critical research area due to its extensive applications in public safety, intelligent surveillance, human-computer interaction, and healthcare monitoring. Accurately understanding human behavior from video sequences is inherently challenging, as it requires simultaneously modeling complex spatial structures and temporal dynamics, while being robust to environmental variations, occlusions, and diverse motion patterns[1].

Traditional deep learning approaches, especially those utilizing 3D CNN, have demonstrated remarkable success in extracting spatiotemporal features by extending 2D convolutional operations into the temporal domain. 3D CNN are capable of capturing local motion patterns and short-term temporal dependencies effectively, contributing significantly to the advancement of video-based recognition systems[2]. However, their inherent limitations in modeling long-range temporal dependencies restrict their ability to fully comprehend behaviors that evolve over extended sequences or that require a broader contextual understanding. This shortcoming often leads to degraded performance in complex real-world scenarios[3].

In contrast, Transformer architectures, originally developed for natural language processing, have recently shown impressive results across various computer vision tasks[4]. Their self-attention mechanism allows for the modeling of global dependencies and long-term interactions among features, enabling a more holistic understanding of temporal and spatial information. Despite these advantages, Transformer-based models typically involve high computational complexity and require large-scale annotated datasets for effective training, making them less practical for resource-constrained applications[5].

Motivated by the complementary strengths of 3D CNN and Transformer, this paper proposes a novel framework that integrates 3D CNN with Transformer modules for video-based behavior recognition. In the proposed framework, 3D CNN serve as the backbone for efficient extraction of local spatiotemporal features, while Transformer components are employed to capture global temporal relationships across video frames[6]. A carefully designed feature fusion strategy ensures seamless integration of local and global information, resulting in a more accurate and robust behavior recognition model without imposing excessive computational overhead.

Through extensive experiments on publicly available benchmark datasets, the proposed framework demonstrates superior performance compared to existing methods, achieving a favorable balance between recognition accuracy and computational efficiency. This work highlights the importance of combining local feature extraction and global temporal modeling, offering a practical and scalable solution for behavior recognition tasks in real-world video analysis systems.

## II. Related Work

**3D CNN in Behavior Recognition**

The application of 3D CNN has significantly advanced the field of video-based behavior recognition. Unlike traditional 2D CNN that only process spatial information from individual frames, 3D CNN extend convolutional operations into the temporal dimension, enabling the simultaneous modeling of spatial and temporal features[7]. By incorporating time as an additional axis, 3D CNN can effectively capture motion cues, short-term dynamics, and contextual variations inherent in video sequences.

Early seminal works, such as C3D, demonstrated the feasibility of applying 3D convolutions to extract generic spatiotemporal features from videos. C3D showed that 3D convolutions could outperform handcrafted feature-based methods by learning task-relevant motion patterns directly from raw video frames. Following this, more advanced architectures like I3D (Inflated 3D ConvNet) further improved performance by inflating 2D filters into 3D, allowing the utilization of pre-trained 2D network while effectively modeling temporal information.

In behavior recognition tasks, 3D CNN have proven to be particularly effective in scenarios involving short and moderate-length actions, such as violent behavior detection, gesture recognition, and activity classification[8]. Their ability to capture localized temporal dependencies makes them well-suited for recognizing actions characterized by distinctive motion patterns over a few frames.

However, despite their strengths, 3D CNN exhibit certain limitations. The receptive field of standard 3D convolutions is often constrained, making it challenging to model long-range temporal dependencies essential for understanding complex or subtle behaviors that evolve over extended periods. Additionally, 3D CNN are computationally intensive, requiring substantial memory and processing power, which can hinder their deployment in real-time or resource-limited environments.

To address these challenges, researchers have explored various strategies, such as incorporating temporal pooling, multi-scale temporal modeling, and two-stream network that separate spatial and temporal processing. Nevertheless, the inherent locality of convolution operations continues to limit the global temporal reasoning capacity of 3D CNN. These limitations motivate the exploration of hybrid approaches, such as integrating Transformer modules, to enhance long-range temporal modeling while maintaining the powerful local feature extraction capabilities of 3D CNN.

**Transformer in Video Understanding**

Transformer, originally proposed for natural language processing tasks, have recently demonstrated remarkable success in computer vision, particularly for video understanding[9]. Unlike convolutional architectures that rely on local receptive fields, Transformer leverage self-attention mechanisms to model global dependencies across spatial and temporal dimensions. This capability is particularly advantageous for video-based tasks, where capturing both short-term dynamics and long-range temporal structures is critical.

Early works such as the TimeSformer and Video Swin Transformer adapted the Transformer architecture to video inputs by introducing mechanisms to efficiently handle the increased computational complexity of spatiotemporal data. TimeSformer, for instance, decomposed attention into spatial and temporal components, enabling scalable learning on video frames. These models showed that Transformer could outperform conventional CNN-based methods on large-scale video action recognition benchmarks, establishing a new paradigm in video understanding research[10].

The key strength of Transformer in video analysis lies in their ability to capture intricate and non-local temporal relationships that are difficult for 3D CNN to model. By attending to all frames simultaneously, Transformer can learn high-level semantic representations that span across an entire video clip, making them particularly effective for recognizing complex actions, subtle motion patterns, and behaviors involving long-term dependencies.

However, the application of pure Transformer architectures in video tasks also faces significant challenges. The quadratic complexity of self-attention with respect to the number of tokens leads to high computational and memory demands, making Transformer less practical for real-time applications or deployment on edge devices. To address this, lightweight variants such as the Video Swin Transformer and Tiny-ViT have been proposed, which introduce hierarchical structures, windowed attention, and efficient tokenization strategies to reduce computational costs while preserving the model's global reasoning capabilities[9].

Transformer have brought transformative changes to video understanding by providing powerful mechanisms for capturing global spatiotemporal contexts[11]. Nevertheless, their computational overhead necessitates careful architectural design, particularly when aiming for real-time performance or deployment in resource-constrained scenarios[12]. These considerations motivate the development of hybrid models that combine the local feature extraction strengths of CNN with the global reasoning abilities of Transformer, aiming to achieve a favorable balance between efficiency and recognition performance.

**3D CNN and Transformer Hybrid Model**

In recent years, hybrid models that combine 3D CNN with Transformer architectures have attracted increasing attention in the field of video understanding. These models aim to integrate the local feature extraction strengths of 3D CNN with the global spatiotemporal modeling capabilities of Transformer, seeking to overcome the limitations inherent in using either architecture alone.

Notable early attempts include adaptations such as SlowFast+Transformer, where the SlowFast network provides multi-pathway temporal feature extraction, and Transformer modules are incorporated to capture long-range dependencies across frames. Similarly, architectures like TimeSformer+CNN hybrids have explored combining convolutional layers for local pattern extraction with Transformer-based self-attention mechanisms for enhanced temporal reasoning. These approaches demonstrated that integrating convolutional inductive biases with the flexibility of attention mechanisms can significantly boost performance in complex action recognition and behavior understanding tasks.

Despite these advancements, many existing hybrid models face challenges related to architectural complexity, high computational cost, and inefficient feature fusion between CNN and Transformer components. Some designs treat CNN and Transformer modules as separate stages, leading to suboptimal information integration, while others introduce heavy overheads that hinder real-time application.

In contrast, the framework proposed in this paper presents a more unified and lightweight fusion strategy. Our method integrates a 3D CNN backbone with a Transformer module in a streamlined manner, allowing for effective spatial-temporal feature interaction without introducing excessive computational burden. Specifically, we leverage the local representation capabilities of 3D convolutions for early-stage feature extraction, followed by Transformer blocks to refine and globally contextualize the temporal dynamics. This tight coupling between 3D CNN and Transformer components enables the model to achieve strong behavior recognition performance while maintaining computational efficiency, making it more practical for real-world deployment scenarios compared to previous hybrid architectures.

## III. Proposed Method

**Overall Architecture**

The proposed framework for video-based behavior recognition is designed to effectively capture both local and global features from the input video sequences by combining a 3D Convolutional Neural Network with Transformer modules. The overall architecture is shown in figure 1.

Initially, the video input is processed through a 3D CNN module. This module consists of eight 3D convolutional layers, each utilizing convolutional kernels of size 3x3x3. The number of kernels in each layer gradually increases as follows: 32, 64, 128, 128, 256, 256, 512, and 512, respectively. This design allows the network to progressively learn more complex features at different levels of abstraction. The 3D convolutional layers are responsible for extracting spatio-temporal features from the video frames, capturing both the spatial and temporal dependencies that are critical for behavior recognition.

Within the 3D CNN module, six 3D pooling layers are applied to reduce the spatial dimensions of the feature maps while preserving essential information. The first pooling layer has a kernel size of 1x2x2, and all subsequent pooling layers use a kernel size of 2x2x2. The pooling operations help lower the computational burden by reducing the size of the feature maps, which is essential for efficient processing in the later stages of the network.

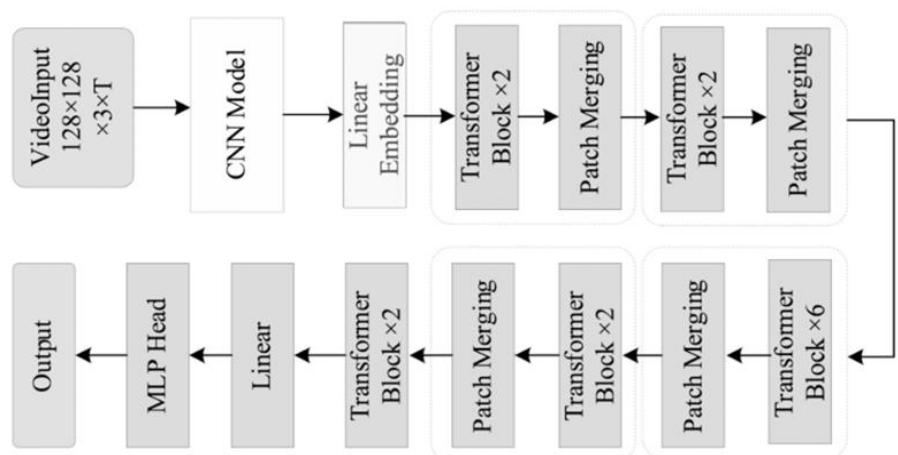

**Figure 1**. The Overall Framework of The Hybrid Model

Once the video input passes through the 3D CNN layers, two Transformer modules are stacked to complete the feature extraction process. Transformers are well-known for their ability to model long-range dependencies and global contextual information. By incorporating these modules after the CNN layers, the model can better capture the broader context of behaviors and recognize actions that involve interactions over time and space across multiple video frames.

Finally, based on the specific requirements of the classification task, appropriate fully connected layers and Multi-Layer Perceptron (MLP) Head modules are added to form the final layers of the network. These layers help to transform the learned features into the desired output for behavior recognition, such as detecting violent actions.

The combination of the 3D CNN and Transformer modules in this framework brings together the best of both architectures. The 3D CNN excels at extracting local spatio-temporal patterns, essential for recognizing actions within individual video frames. In contrast, the Transformer modules are adept at capturing long-range dependencies and contextual information across frames. This synergy allows the hybrid model to handle the complexity of video-based behavior recognition, improving its accuracy in detecting violent behaviors despite challenges such as background noise, occlusions, and varying video resolutions. By leveraging both the local feature extraction capabilities of the CNN and the global feature representation of the Transformer, the model achieves a robust and efficient solution for video-based behavior recognition.

**3D CNN-Based Spatiotemporal Feature Extraction**

The 3D CNN module is designed to effectively extract spatio-temporal features from video sequences, crucial for behavior recognition tasks. The module consists of eight 3D convolutional layers, each using 3x3x3 convolutional kernels. These kernels are particularly suited for processing video data as they capture local spatial patterns within individual frames as well as temporal dependencies across consecutive frames. The number of kernels in the layers increases progressively: 32, 64, 128, 128, 256, 256, 512, and 512. This incremental growth enables the network to learn increasingly complex features as the data passes through the layers, starting from simple patterns like edges and textures to more abstract representations like motion and interactions. The module also includes six 3D pooling layers, with the first pooling layer using a kernel size of 1x2x2, and the remaining layers using a 2x2x2 kernel. The pooling layers reduce the spatial and temporal dimensions of the feature maps, preserving the most informative features while lowering the computational load. The convolution operation can be mathematically expressed as:

$$F_{out(i,j,k)} = \sum_{m=1}^{M} \sum_{n=1}^{N} \sum_{p=1}^{P} F_{in}(i+m, j+n, k+p) \cdot W(m,n,p)$$

where $F_{out(i,j,k)}$ represents the input feature map at position (i,j,k) and $W(m,n,p)$ is the 3D kernel. This 3D convolution operation allows the network to learn both spatial and temporal features, essential for recognizing behavior patterns over time. The 3D CNN module's design enables the model to capture local spatio-temporal patterns effectively, making it well-suited for recognizing complex behaviors that involve motion and interactions across multiple frames.

**Transformer-Based Temporal Modeling**

After the 3D CNN module extracts the spatio-temporal features from the video input, these features are passed into Transformer modules for enhanced temporal modeling. Transformers are well-known for their ability to capture long-range dependencies and global contextual information, making them ideal for processing sequential data such as video frames. The input to the Transformer module is the feature map produced by the 3D CNN, which is reshaped into a sequence format. Specifically, each spatial-temporal feature map is flattened into a one-dimensional vector, and each of these vectors corresponds to a token in the sequence. These tokens represent different regions of the video and contain information about the spatial and temporal features learned by the CNN. To help the Transformer understand the position of each token within the sequence, position encodings are added to the input tokens. These position encodings provide information about the relative or absolute position of the tokens within the sequence, allowing the Transformer to differentiate between tokens at different times or spatial locations. The position encoding is typically added to the input feature map, and it can be learned or fixed.

The attention mechanism within the Transformer then processes the sequence of tokens by computing self-attention, which enables the model to weigh the importance of each token in relation to others in the sequence. This attention mechanism is crucial for capturing long-range dependencies across frames, allowing the model to understand how different frames in the video interact with one another over time. The self-attention operation can be mathematically expressed as:

$$Attention(Q, K, V) = softmax\left(\frac{QK^T}{\sqrt{d_k}}\right)V$$

Where $Q$ represents the query matrix, $K$ is the key matrix, $V$ is the value matrix, $d_k$ is the dimensionality of the query and key vectors.

This formula calculates the attention scores by taking the dot product of the query and key matrices, scaled by the square root of the key dimensionality $\sqrt{d_k}$ followed by a softmax operation to obtain the attention weights. The weighted sum of the value matrix $V$ is then computed to produce the output of the attention mechanism.

The self-attention process computes the attention score for each token pair, which reflects their relative importance, and generates a weighted sum of the tokens for each output. This enables the model to focus on key temporal patterns and interactions across the video frames, improving its ability to recognize complex behaviors that span multiple frames or involve interactions between different entities. By incorporating the Transformer, the model effectively combines local spatio-temporal features learned by the 3D CNN with global temporal context, providing a powerful framework for video-based behavior recognition.

**Fusion Strategy**

To effectively integrate the features extracted by the 3D CNN module with the output from the Transformer, a fusion strategy is employed that combines both local spatio-temporal information and global temporal context. The key challenge in this fusion is to ensure that the distinct characteristics of both models are preserved while enabling the model to learn complementary representations from both types of features. In this work, we adopt a weighted fusion approach, where the features from the 3D CNN and the Transformer are combined through a learned weight mechanism. This strategy allows the model to dynamically assign importance to each type of feature based on the context of the video data. Specifically, the output from the 3D CNN is first flattened and passed through a series of fully connected layers to match the dimensionality of the Transformer output. Then, a weighted sum of the two feature vectors is computed, where the weights are learned during training. This can be expressed mathematically as:

$$F_{fused} = \alpha \cdot F_{CNN} + (1 - \alpha) F_{Transformer}$$

Where $F_{CNN}$ and $F_{Transformer}$ represent the feature vectors from the 3D CNN and the Transformer, respectively, and $\alpha$ is a learned scalar weight that controls the contribution of each feature set. This approach enables the model to adjust the importance of each feature type based on the behavior being recognized, allowing the system to focus more on local spatio-temporal patterns when needed and more on global temporal dependencies in other situations. Additionally, residual connections are introduced to facilitate gradient flow and ensure the model retains important low-level features from the 3D CNN. The final fused feature representation is then passed through fully connected layers and an MLP Head to output the final detection results. By employing this fusion strategy, the model combines the strengths of both architectures—CNN's ability to capture local motion patterns and Transformer's power to model long-range temporal dependencies—resulting in improved accuracy and robustness in video-based behavior recognition tasks.

## IV. Experiments And Results

**Datasets**

For the evaluation of our proposed framework, we utilize several publicly available datasets commonly used in video-based behavior recognition tasks. These datasets provide a diverse set of video sequences that involve various actions and behaviors, allowing us to assess the performance of the model across different scenarios. One of the primary datasets used is the Hockey Fight dataset, which consists of videos depicting fights in hockey games. This dataset contains annotated frames showing violent actions, making it an ideal choice for evaluating the model's ability to recognize aggressive behavior. The videos in this dataset vary in length, ranging from a few seconds to several minutes, and contain different numbers of players, adding complexity to the recognition task. The original resolution of the videos is typically 1920x1080 pixels, but for consistent processing, all frames are resized and cropped to a resolution of 224x224 pixels.

Another key dataset used is the RWF-2000 dataset, which is a large-scale dataset for recognizing real-world fighting behaviors. It contains 2,000 video clips, each labeled with one of several action categories, such as fighting, hugging, and others. The videos in this dataset also vary in resolution, typically 1920x1080 pixels, and, similar to the Hockey Fight dataset, all frames are resized and cropped to 224x224 pixels to maintain consistency across the input data. These datasets are chosen due to their wide use in behavior recognition research and the clear annotations they provide for ground-truth comparison. The diverse nature of these datasets ensures that our model is rigorously tested, not only on specific types of violent behaviors but also on general behavior recognition tasks, offering a comprehensive evaluation of its performance.

**Experimental Environment**

The experiments were conducted on a system equipped with an NVIDIA RTX 3060 GPU, which features 24 GB of VRAM. This powerful GPU was chosen for its ability to efficiently handle the computational demands of training and inference in deep learning models. Its parallel processing capabilities significantly accelerated the computations, making it an ideal choice for the intensive tasks involved in video-based behavior recognition. PyTorch 2.0.1 was utilized as the primary deep learning framework, selected for its dynamic computation graph and robust support for GPU acceleration, which facilitated faster model development and experimentation. The environment was configured to ensure that all experiments were performed in a controlled setting with standardized computational resources, thus ensuring the reproducibility of results. To maintain consistency and fairness during both training and evaluation, all datasets were split into training and testing sets with an approximate ratio of 8:2. For preprocessing, all video frames were resized to a uniform resolution of 224×224 pixels. Additionally, standard data augmentation techniques, such as random cropping and horizontal flipping, were employed to enhance the model's generalization ability and prevent overfitting. The configuration of the experimental environment is summarized in Table 1.

**Table 1. Experimental Environment Configuration**

| Parameter | Configuration |
|---|---|
| GPU | NVIDIA RTX 3060, 24 GB VRAM |
| CPU | 6 Core, Intel i5-12600 |
| RAM | 16GB |
| Memory | 512GB |
| CUDA | 12.4 |
| cuDNN | 8.3.3 |
| Python | 3.10.12 |
| PyTorch | 2.0.1 |
| Input Frame Size | 224*224 |
| Number of Epochs | 200 |
| Batch Size | 32 |
| Learning Rate | 0.001 |
| Optimizer | Adamw |
| Loss Function | Cross-Entropy Loss |
| Normalization Mean | [0.485, 0.456, 0.406] |
| Normalization Std | [0.229, 0.224, 0.225] |

**Results and Analysis**
**Evaluation of Loss and Accuracy**

Throughout the training process, the model is optimized using the Adam optimizer with an initial learning rate of 1e-4. To ensure efficient convergence, a Learning Rate Scheduler is employed to adjust the learning rate dynamically. Specifically, the learning rate is reduced by a factor of 0.1 when the Validation Loss plateaus for 5 consecutive epochs. This gradual decrease in the learning rate encourages the model to settle into a better local minimum, thereby improving the overall performance of the model. The training is performed on a high-performance GPU, which accelerates the computations, and mixed precision training is used to optimize memory usage and speed up training.

To monitor the progress of model convergence and its ability to generalize, both the Training Loss and Validation Loss are tracked continuously throughout the training process. This systematic approach helps ensure that the 3D CNN + Transformer model achieves optimal performance while avoiding overfitting. The results of the evaluation on the Hockey Fight and RWF-2000 datasets are presented by analyzing the loss curves and accuracy metrics, as shown in figures 2 and 3. On the Hockey Fight dataset, the 3D CNN + Transformer model achieves an impressive accuracy of 96.7%, highlighting the effectiveness of combining the spatial feature extraction power of the 3D CNN with the long-range dependency modeling capabilities of the Transformer. On the more complex and diverse RWF-2000 dataset, the model maintains a consistent performance trend, outperforming other models with an accuracy of 93.56%. This demonstrates the model's strong generalization ability across different types of behavior recognition tasks. The steady loss reduction and high accuracy on both datasets emphasize the robustness and adaptability of the 3D CNN + Transformer hybrid architecture.

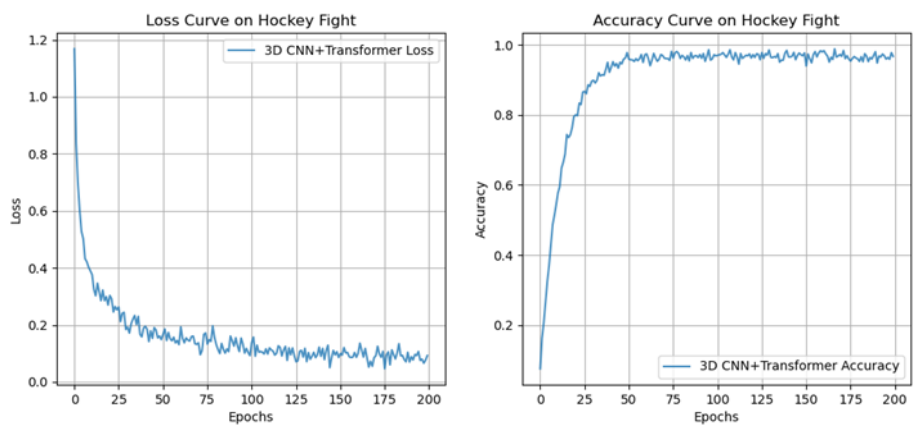

**Figure 2. Loss and Accuracy Curve on Hockey Fight Dataset**

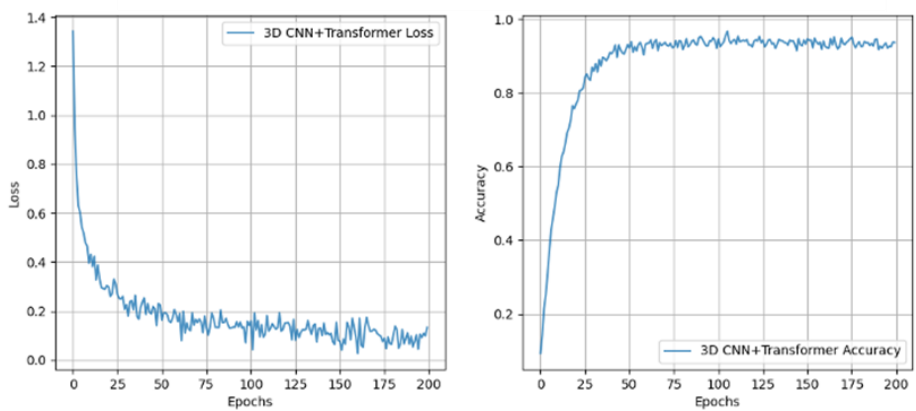

**Figure 3. Loss and Accuracy Curve on RWF-2000 Dataset**

**Performance Comparison with Other Models**

In this experiment, we evaluate the performance of three network architectures—LSTM, 3D CNN, and the hybrid 3D CNN+Transformer—on the Hockey Fight and RWF-2000 datasets. The objective is to assess the effectiveness of each model in learning and generalizing for violence detection tasks.

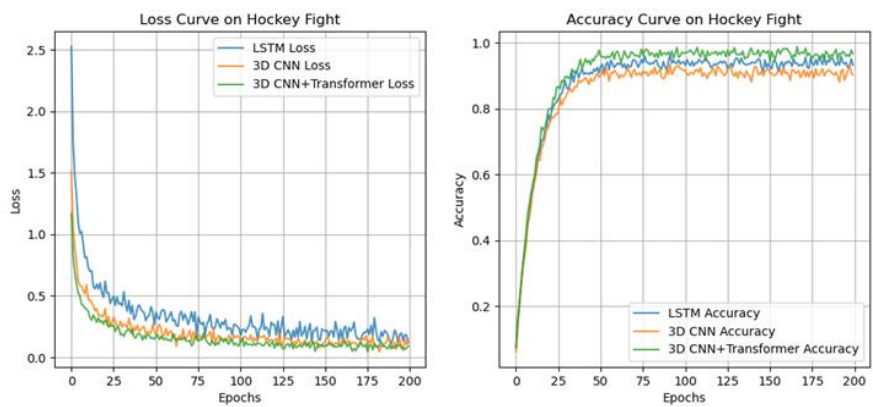

**Figure 4. Loss Curve Comparison on Hockey Fight Dataset**

As shown in figure 4, on the Hockey Fight dataset, the 3D CNN+Transformer model achieves the highest accuracy of 96.7%, highlighting the advantages of combining the spatial feature extraction power of 3D CNN with the long-range dependency modeling capability of Transformers. The LSTM model attains an accuracy of 93.9%, demonstrating that sequential modeling can capture temporal dependencies effectively, but it lacks the robust spatial feature extraction capabilities of CNN. Meanwhile, the standalone 3D CNN model achieves 91%, which, although leveraging spatiotemporal features, shows limitations in capturing long-range dependencies when compared to the hybrid model.

For the RWF-2000 dataset, shown in figure 5, the performance trend remains consistent. The 3D CNN+Transformer model again outperforms the others, achieving an accuracy of 93.56%, which underscores its strong generalization ability. The 3D CNN model achieves 92.07%, benefiting from spatiotemporal feature extraction, but still lags behind the hybrid model due to its limited ability to capture global features. The LSTM model, while effective in sequential learning, achieves 89.03%, reflecting its relative difficulty in handling the complex spatial patterns present in crowd violence scenarios.

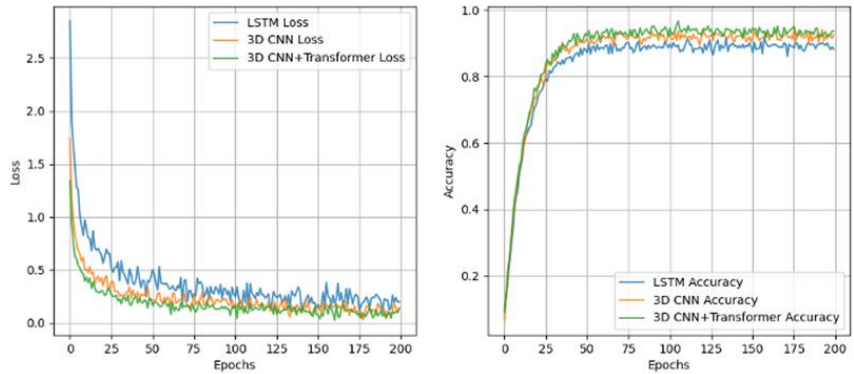

**Figure 5. Accuracy Curve Comparison on RWF-2000 Dataset**

The loss curves for all models further demonstrate that the 3D CNN+Transformer architecture converges more smoothly and stabilizes earlier than the other two models, suggesting enhanced learning efficiency and reduced overfitting. In contrast, the LSTM model exhibits more fluctuations in both loss and accuracy, potentially due to its reliance on sequential feature extraction without explicit spatial representations. The 3D CNN model, although more stable, shows a slightly slower convergence rate compared to the hybrid model, emphasizing the importance of integrating global feature extraction from Transformers to achieve faster and more efficient convergence.

**ROC-AUC Comparison**

The AUC (Area Under the Curve) of the ROC (Receiver Operating Characteristic) curve provides a quantitative measure of a classifier's ability to distinguish between violent and non-violent actions. On the Hockey Fight dataset, shown in figure 6, the 3D CNN+Transformer model achieves an AUC of 0.9652, significantly outperforming traditional handcrafted feature-based methods, such as ViF+OViF, which achieves an AUC of 0.9281. This substantial improvement highlights the superior ability of the hybrid model to capture complex spatiotemporal patterns, thus enhancing classification performance.

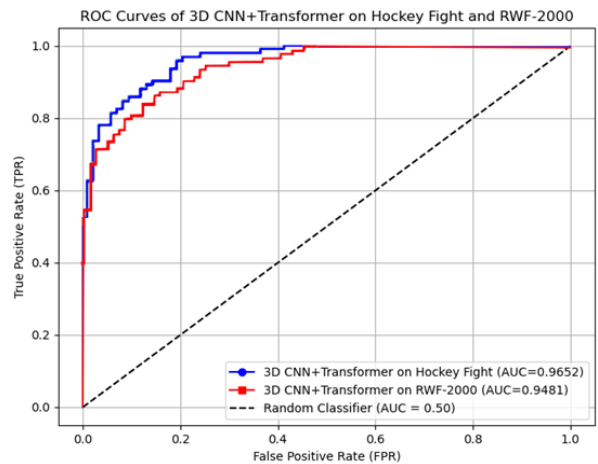

**Figure 6. ROC-AUC Curve**

For the RWF-2000 dataset, the 3D CNN+Transformer model attains an AUC of 0.9481. While slightly lower than that on the Hockey Fight dataset, this performance is still impressive considering the inherent challenges of the RWF-2000 dataset, including complex backgrounds, a larger number of individuals, and lower video resolution. The modest decrease in AUC is expected due to these factors; nevertheless, the model maintains strong detection performance, demonstrating its robust generalization ability across various violent scenarios. This reinforces the effectiveness of the 3D CNN+Transformer architecture in tackling diverse and complex behavior recognition tasks.

## V. Conclusion

In this study, we presented a novel hybrid model that combines 3D CNN and Transformers for video-based behavior recognition, specifically targeting violent behavior detection. The proposed architecture successfully integrates the spatial feature extraction power of 3D CNN with the long-range dependency modeling capabilities of Transformers, offering a comprehensive solution for recognizing complex behaviors in video sequences.

The experimental results on the Hockey Fight and RWF-2000 datasets show that the 3D CNN+Transformer model outperforms traditional models, such as standalone 3D CNN and LSTM network, in terms of both accuracy and AUC scores. This demonstrates the superiority of the hybrid model in handling the challenges of spatiotemporal data and its ability to generalize effectively across different scenarios. The model's robustness to varying conditions, such as background complexity and video resolution, further supports its practical applicability in real-world settings.

Future work can explore further optimizations of the model, such as incorporating additional attention mechanisms or integrating multi-modal data for enhanced performance. Additionally, the model can be adapted for other complex behavior recognition tasks, including monitoring in public spaces, surveillance, and safety systems.

In conclusion, the proposed 3D CNN+Transformer hybrid model provides a promising approach for video-based behavior recognition, achieving high performance on diverse datasets and demonstrating strong potential for practical deployment in behavior detection applications.

## References

[1] Accattoli, S., Et Al., Violence Detection In Videos By Combining 3D Convolutional Neural Networks And Support Vector Machines. Applied Artificial Intelligence, 2020. 34(4): P. 329-344.
[2] Appavu, N. Violence Detection Based On Multisource Deep CNN With Handcraft Features. In 2023 IEEE International Conference On Advanced Systems And Emergent Technologies (IC_ASET). 2023. IEEE.
[3] Dong, J., Et Al., Action Recognition Based On The Fusion Of Graph Convolutional Networks With High Order Features. Applied Sciences, 2020. 10(4): P. 1482.
[4] Cheng, M., K. Cai, And M. Li. RWF-2000: An Open Large Scale Video Database For Violence Detection. In 2020 25th International Conference On Pattern Recognition (ICPR). 2021. IEEE.
[5] García-Gómez, J., Et Al. Violence Detection In Real Environments For Smart Cities. In Ubiquitous Computing And Ambient Intelligence: 10th International Conference, Ucami 2016, San Bartolomé De Tirajana, Gran Canaria, Spain, November 29–December 2, 2016, Part II 10. 2016. Springer.
[6] Hassner, T., Y. Itcher, And O. Kliper-Gross. Violent Flows: Real-Time Detection Of Violent Crowd Behavior. In 2012 IEEE Computer Society Conference On Computer Vision And Pattern Recognition Workshops. 2012. IEEE.
[7] Chen, J., Et Al. An Improved Two-Stream 3D Convolutional Neural Network For Human Action Recognition. In 2019 25th International Conference On Automation And Computing (ICAC). 2019. IEEE.
[8] Ji, S., Et Al., 3D Convolutional Neural Networks For Human Action Recognition. IEEE Transactions On Pattern Analysis And Machine Intelligence, 2012. 35(1): P. 221-231.
[9] Liu, Z., Et Al. Video Swin Transformer. In Proceedings Of The IEEE/CVF Conference On Computer Vision And Pattern Recognition. 2022.
[10] Liu, Z., Et Al. Swin Transformer: Hierarchical Vision Transformer Using Shifted Windows. In Proceedings Of The IEEE/CVF International Conference On Computer Vision. 2021.
[11] Ulhaq, A., Et Al., Vision Transformers For Action Recognition: A Survey. Arxiv Preprint Arxiv:2209.05700, 2022.
[12] Tran, D., Et Al. Learning Spatiotemporal Features With 3d Convolutional Networks. In Proceedings Of The IEEE International Conference On Computer Vision. 2015.